\documentclass[runningheads]{llncs}
\usepackage{graphicx}
\usepackage{caption}
\usepackage{xcolor}
\usepackage{siunitx}
\usepackage{mathabx}
\usepackage{cite}
\begin{document}
\title{AI Playground: Unreal Engine-based Data Ablation Tool for Deep Learning}
%
%
\author{Mehdi Mousavi\inst{1}\orcidID{0000-0001-8948-8011} \and Aashis Khanal\inst{1}\orcidID{0000-0002-0164-2465} \and
Rolando Estrada\inst{1}\orcidID{0000-0003-1607-2618}}
\authorrunning{M. Mousavi et al.}
%
\institute{Department of Computer Science, Georgia State University, Atlanta GA 30303, USA
\email{\{smousavi2,akhanal1\}@student.gsu.com}, \email{restrada1@gsu.edu}}
\maketitle              

\begin{abstract}
Machine learning requires data, but acquiring and labeling real-world data is challenging, expensive, and time-consuming. More importantly, it is nearly impossible to alter real data post-acquisition (e.g., change the illumination of a room), making it very difficult to measure how specific properties of the data affect performance. In this paper, we present AI Playground (AIP), an open-source, Unreal Engine-based tool for generating and labeling virtual image data. With AIP, it is trivial to capture the same image under different conditions (e.g., fidelity, lighting, etc.) and with different ground truths (e.g., depth or surface normal values). AIP is easily extendable and can be used with or without code. To validate our proposed tool, we generated eight datasets of otherwise identical but varying lighting and fidelity conditions. We then trained deep neural networks to predict (1) depth values, (2) surface normals, or (3) object labels and assessed each network's intra- and cross-dataset performance. Among other insights, we verified that sensitivity to different settings is problem-dependent. We confirmed the findings of other studies that segmentation models are very sensitive to fidelity, but we also found that they are just as sensitive to lighting. In contrast, depth and normal estimation models seem to be less sensitive to fidelity or lighting and more sensitive to the structure of the image. Finally, we tested our trained depth-estimation networks on two real-world datasets and obtained results comparable to training on real data alone, confirming that our virtual environments are realistic enough for real-world tasks. 
\keywords{Synthetic data  \and Deep learning \and Virtual environment}
\end{abstract}

\begin{figure}[htbp]
\centerline{\includegraphics[width = 0.73\textwidth]{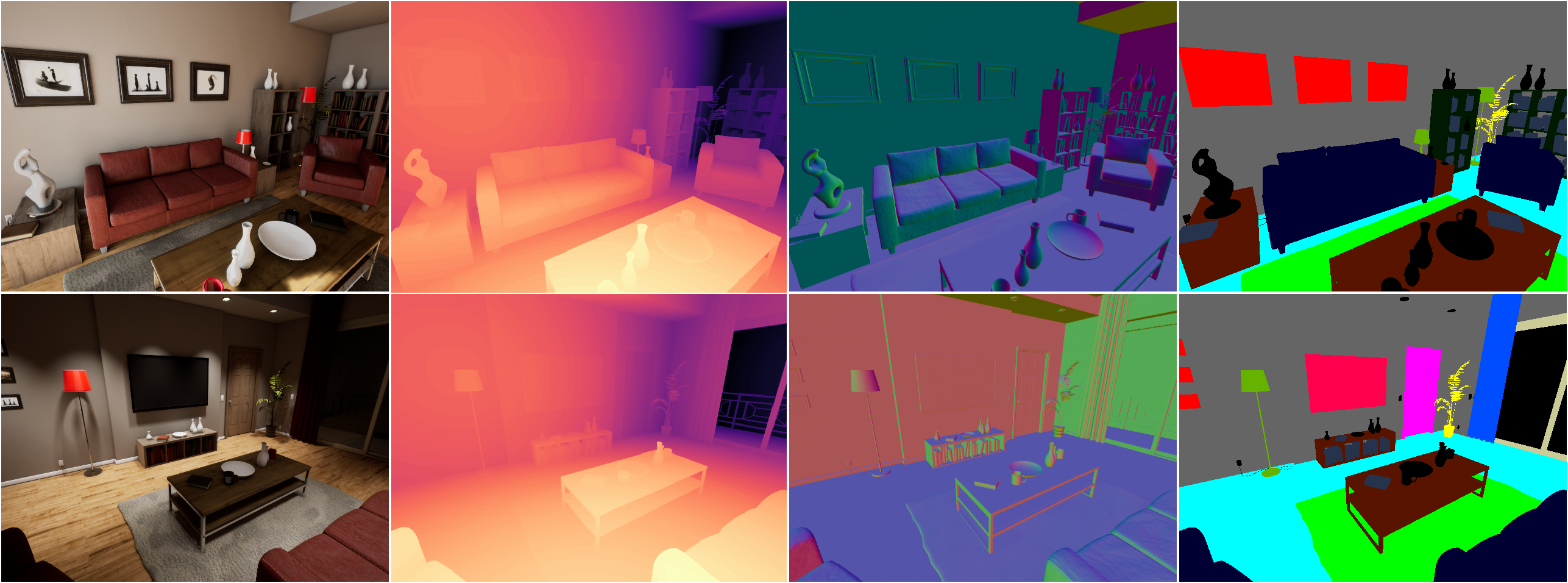}}
\caption{\textbf{Virtual environments:} Sample screenshots from our annotated virtual environments. From left to right: depth, surface normals, and semantic labels}
\label{fig:envexamples}
\vspace{-1em}
\end{figure}

\section{Introduction}

The remarkable success of deep learning in recent years would not have been possible without large, high-quality datasets \cite{LeCun2015}. Deep neural networks have thousands or even millions of parameters, which require vast numbers of training examples to tune. However, producing a high-quality dataset of real data is very challenging. First, one has to acquire the raw data, an often laborious task. Second, the training data must either be labeled manually---which is slow, subjective and may require significant expertise---or with expensive, specialized equipment. Finally, errors can occur in both the acquisition and labeling phases.



 

Real data also has an additional, more subtle limitation: it is very difficult to control before acquisition and nearly impossible to change afterwards. For instance, once an image has been taken, one cannot change its illumination from day to night or replace one object for another\footnote{Photo-manipulation techniques can be used to alter images, but their effects are either non-specific (e.g., reducing brightness) or introduce unwanted artifacts. They also require significant human effort.}. The only way to achieve these effects is by manipulating the source of the data before acquisition; however, this approach requires a controlled environment and precise measurements. For example, to change the color of a couch one would need to swap out two otherwise identical couches and place them in the same, exact location. Aside from its difficulty, this approach is not feasible for natural scenes or crowd-sourced data.

The above limitation makes it is very difficult to isolate the impact of individual features on a system's performance. For example, imagine that we want to assess how an object's texture affects our system's ability to segment it. In this case, we would need to compare our system's output across different objects and hope that the impact of other features, e.g., lighting or shape, cancels out across the samples. As such, data ablation studies are rare in machine learning. Most ablation analyses add/remove either (1) components of the model \cite{2019arXiv190108644M} or (2) secondary features computed from the data \cite{2019arXiv191000174M}. The latter is close in spirit to data ablation but is more limited, since secondary features are dependent on the raw, unchangeable data.


\begin{figure}[htbp]
\centerline{\includegraphics[width = 0.83\textwidth]{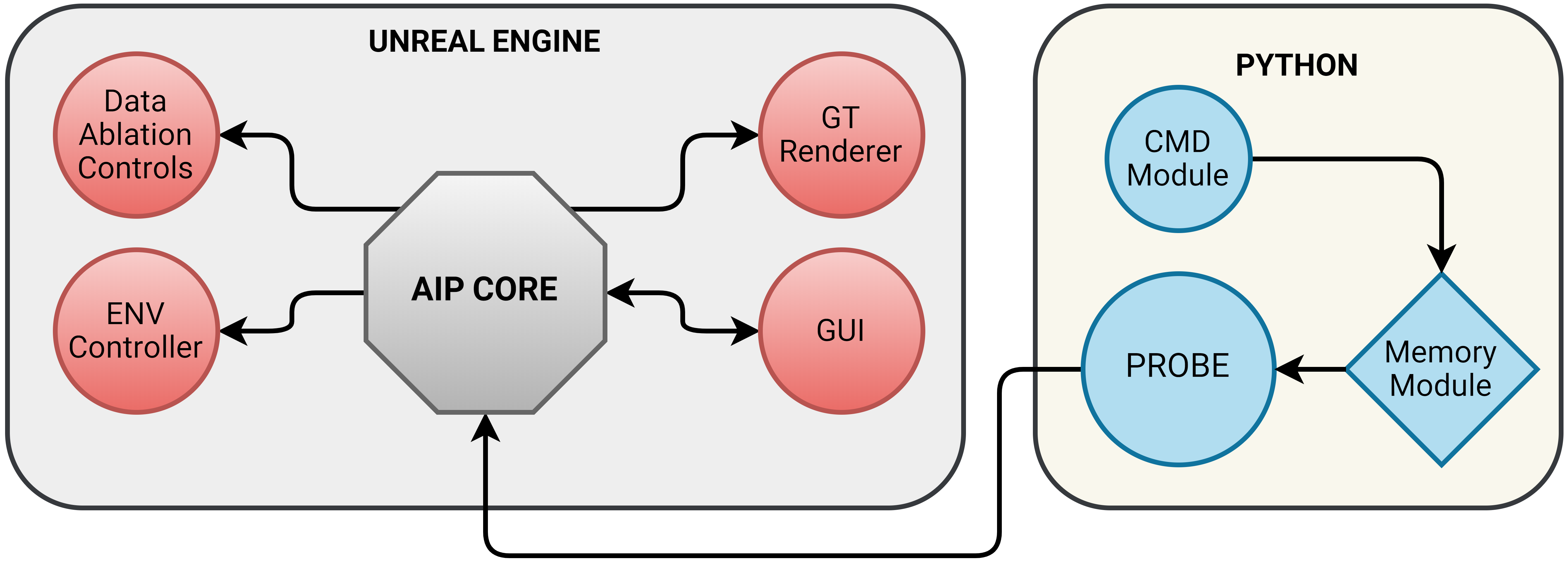}}
\caption{\textbf{AI Playground:} Our tool has two main modules: the \textit{AIP Core} within UE4 and \textit{Probe}, a Python module that communicates with the Core. Probe receives instructions generated by the Command module, and saves its state in its own dedicated memory. This allows changing settings inside the engine while AIP is running. Manually changing components is also possible via the GUI.}
\label{fig:AIPStruct}
\vspace{-1em}
\end{figure}

To address this gap, we developed AI Playground, a user-friendly tool based on the Unreal Engine (Epic Games, USA) \cite{unrealengine} that supports data ablation studies in computer vision.\footnote{Source code, documentation, supplementary images, and high-definition figures are available on our GitHub page: \url{https://git.io/JJkhQ}} Our system allows researchers to easily capture synthetic data from fully customizable virtual environments; this data can then be used to train or test an AI system. Virtual data is free from acquisition errors or labeling bias and is ideal for the data ablation studies discussed above, e.g., capturing the same image under multiple lighting conditions. More importantly, as our experiments confirm, today's high-resolution computer graphics are realistic enough to be used for training deep neural networks on real-world tasks.


As we detail in Sec.~\ref{sec:methodology}, AI Playground is an open-source UE project with four main components: (1) a set of high-resolution environments; (2) multiple ground-truth annotations (e.g., depth, surface normals, etc.); (3) built-in tools for data ablation (e.g., for adjusting lighting, polygon resolution, etc.); and (4) a user-friendly, graphical interface. Users can either run our system as a pre-built application or import it as a regular UE project. In the latter case, users can extend their local version of AIPlayground or copy parts of it (e.g., scripts) for use in their own projects. It is easy to add custom environments or ground-truth annotations without writing any code. And we provide sample code and the necessary documentation to add new forms of data ablation to AIPlayground. Figure~\ref{fig:AIPStruct} provides a flowchart of our tool.





To validate its usefulness, we used AIP to carry out a series of data ablation studies. As detailed in Sec.~\ref{sec:experiments}, we trained and tested deep networks on (1) monocular depth estimation, (2) surface normal estimation, and (3) semantic segmentation. AIP allowed us to draw novel insights about feature importance (Sec.~\ref{sec6:Discussion}), and we also confirmed that networks trained on depth estimation via AI Playground achieve good performance on real-world datasets.

\section{Related Work}
\label{sec:relatedWork}
Data-hungry models like DCNN (Deep Convolutional Neural Networks) have generated newfound interest in virtual data \cite{Photorealism2,photorealism3,datafromgames}. One popular approach is to use modded old video games (e.g., Atari games \cite{2013arXiv1312.5602M}). However, this approach lacks customizability and photo-realism. This data cannot be customized to fit a more specific problem and using old video games introduces a lack of photo-realism that has been proven beneficial for virtual data\cite{7926706_photo_realism, 2019arXiv190108644M}. In contrast, Veeravasarapu \textit{et al.} \cite{7926706_photo_realism} used probabilistic generative models to create random environments in Blender (Blender Foundation, The Netherlands) \cite{blender}. However, these probabilistic models need to be manually adapted for each type of desired environment. For example, the probabilistic model of an outdoor street scene varies significantly from one of an interior environment. Also, while randomness is useful for quickly creating novel environments, these environments may not be faithful to reality. For example, a random probabilistic model might decide to put a couch on a table, which never happens in the real world. Furthermore, depending on hardware, rendering an image in Blender using ray-tracing can take up to a minute or more; the same level of fidelity can be achieved in game engines in real-time. As mentioned in \cite{7926706_photo_realism}, generating a Path-traced image in Blender takes up to 9 minutes (547 seconds), and ray-tracing based rendering for a single image can take 20 seconds or more.

In another study, researchers used 3D reconstruction to generate a photo-realistic 3D scene that allows limited interaction such as walking around \cite{habitat19iccv}. This method requires expensive equipment and complex calculations to generate the pixel-wise ground-truth for tasks like \textit{depth estimation} and \textit{surface normal estimation}. The generated ground-truth and 3D environment are subject to artifacts and estimation errors appearing as black spots in the images. Also, these environments are extremely hard to expand as they require costly specialized equipment for measurement. 

 
The work most similar to our proposed system is UnrealCV---an Unreal Engine 4 (UE4) plugin that has been used in a number of research projects. UnrealCV provides an interface to communicate with the Unreal Engine for computer vision and robotics research \cite{qiu2017unrealcv}. However, UnrealCV requires command-line-based interaction and C++ coding. As such, it has a high barrier of entry and can be discouraging for computer vision researchers who are unfamiliar with game engines. It also lacks intuitive dials and knobs for dynamic interaction with the environment. More importantly, it is not built for data ablation; any systematic changes in fidelity, lighting, etc. have to be coded from scratch by the researcher.        

In contrast, our goal is to reduce the skill level need to obtain virtual pixel-perfect data. Our approach is accessible, user-friendly, and has many intuitive ways to interact with the environment. We use the high quality renderer integrated in Unreal Engine to produce lifelike synthetic images, and AIP does not require any knowledge of UE4 programming. As we detail in the following section, our companion Python module (Probe) communicates with the UE4 application to control the environment and take samples while keeping a record of every step for image re-creation. 



\begin{figure*}[h]
\centerline{\includegraphics[width = 0.73\textwidth]{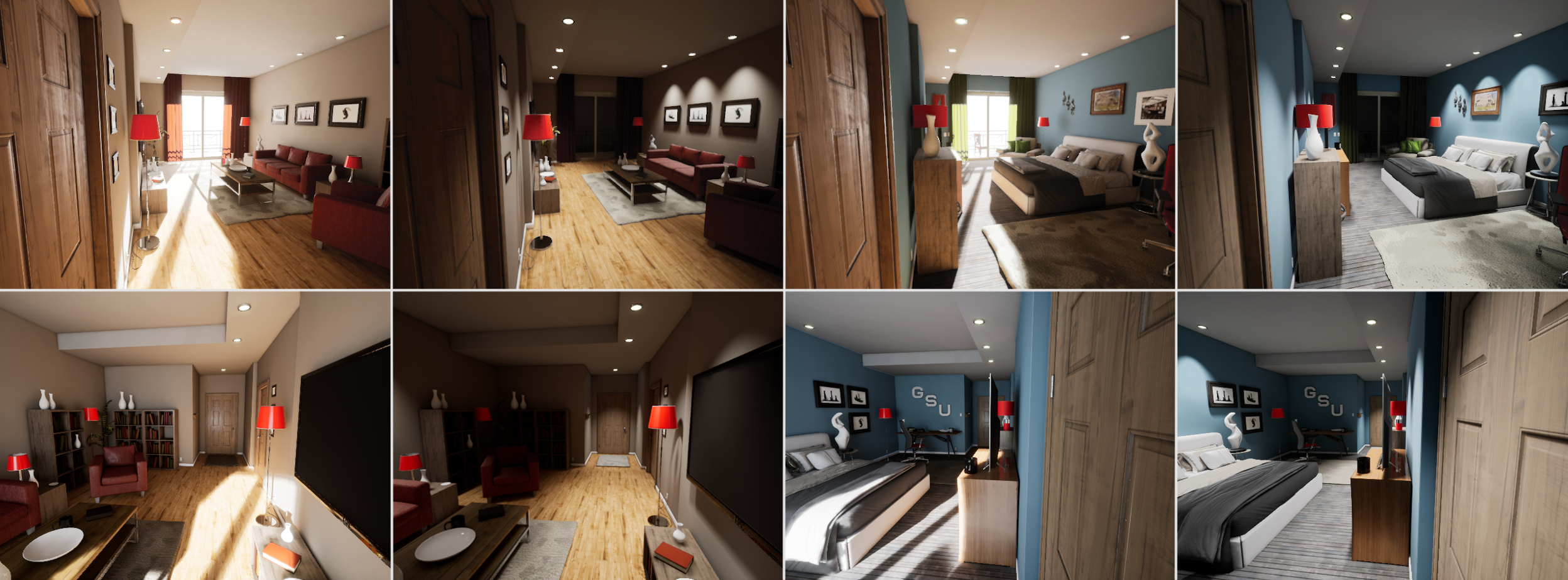}}
\caption{Sample images captured by Probe. Left to right:  Brown Room Day, Brown Room Night, Blue Room Day, Blue Room Night (All high settings)}
\label{fig:scenarios}
\vspace{-1em}
\end{figure*}

\section{AIPlayground}
\label{sec:methodology}
AIPlayground is a UE4-based tool for data ablation studies in computer vision. Unreal Engine is the engine of choice for video games with high-resolution, real-time 3D graphics. It is free for both commercial and non-commercial use and its source code is publicly available (though not fully open source). As illustrated in Fig.~\ref{fig:AIPStruct}, our system has four components: (1) high-resolution 3D environments; (2) multiple ground-truth annotations; (3) data ablation controls; and (4) a user-friendly, graphical interface. As we discuss further below, we use Blueprint, Unreal's visual scripting language, for the ground-truth annotations and data ablation controls. We use a separate Python interaction module|Probe|for data collection, which is also publicly available. 



\subsection{Three-dimensional environments}
In addition to being a game engine, Unreal Engine provides powerful tools for realistic architectural visualization. As such, we developed two environments based on UE4's built-in "Realistic Rendering" scene, dubbed Brown Room and Blue Room in our experiments. Each environment has two general lighting profiles, Day and Night, as illustrated in Fig.~\ref{fig:scenarios}. To mimic existing real-world datasets, the environments are static (i.e., no movement of the components aside from the probe character). AIP currently uses static (i.e., baked) lighting to illuminate the scene. Baking light-maps is a commonly used method to simulate high-fidelity lighting on lower-capacity hardware. It uses ray-tracing to determine dark and light spots in the scene and paints the textures on those areas to look accordingly. The result is a very realistic environment that is rendered rapidly with little to no extra computation required at run-time. This means AIP supports very high frame-rates, which allow for fast data acquisition. We can switch between different ground-truth annotations in fractions of a second without causing artifacts such as blur, fuzziness on the edges, or motion-blur. 
\begin{figure}[htbp]
\centerline{\includegraphics[width = 0.73\textwidth]{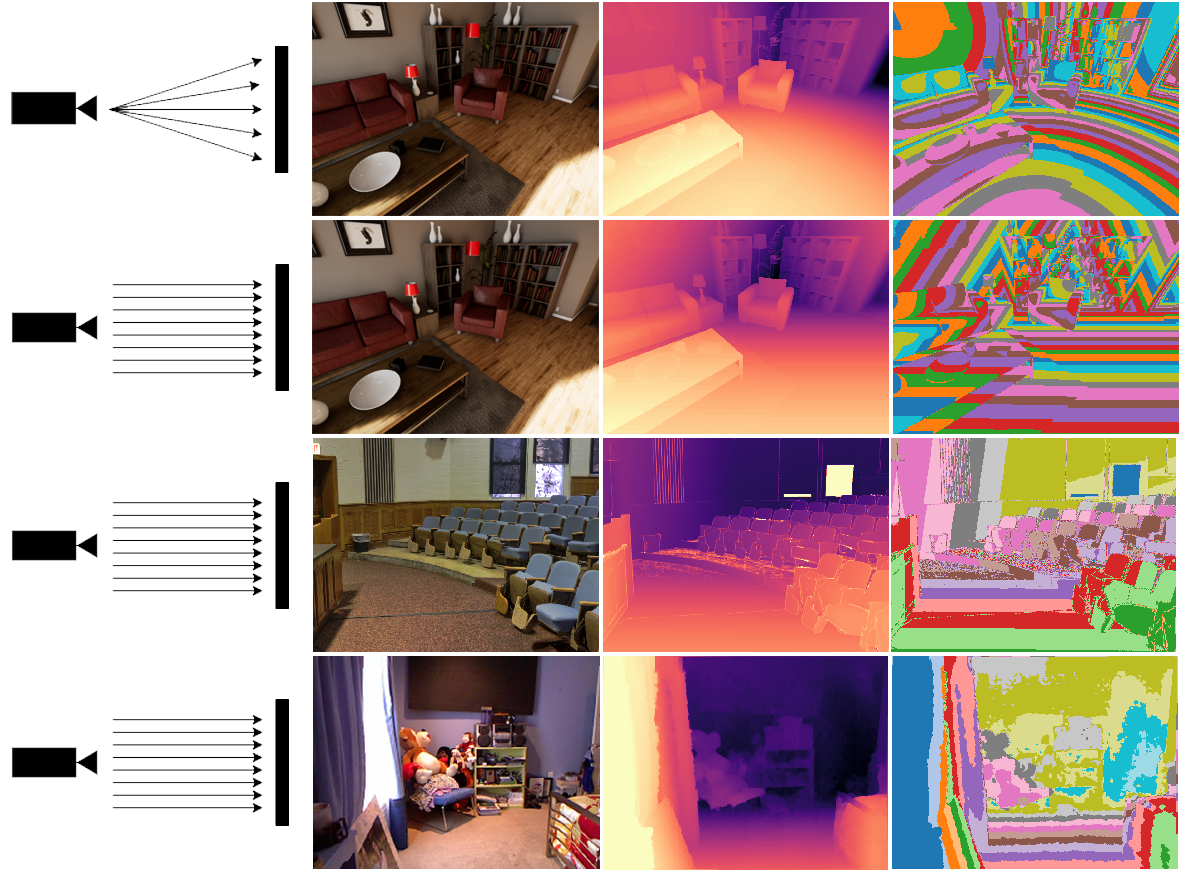}}
\caption{\textbf{Depth estimation:} AIP uses perspective projection (first row), which is more accurate than orthographic projection (second row). The third column uses color banding to highlight the differences between these two approaches. The bottom rows show examples from the DIODE and NYUv2 datasets. Note the lack of artifacts in the virtual ground truth.}
\label{fig:prespvortho}
\vspace{-1em}
\end{figure}

\subsection{Ground-truth annotations}
One of the main advantages of virtual environments is that obtaining ground-truth annotations is trivial relative to real-world environments. Specifically, we use Unreal Blueprint (an internal scripting language) to calculate the ground-truth properties listed below. AIP includes Blueprint scripts for estimating depth, surface normals, and object classes, and can be readily extended by adding additional scripts. We use post-processing shaders, called materials in UE4, to overlay these properties over the image, enabling pixel-perfect alignment between the data and the ground-truth labels (see Fig.~\ref{fig:envexamples} for examples).
\vspace{0.4em}

\noindent \textbf{Depth estimation:} We calculate the normalized distance between each pixel that belongs to a specific object and the camera. We set the real-life range of depth to 10 meters, which covers the entire environment and does not clip between any corners of the room. We define the depth using \textit{perspective projection} relative to the viewer's POV, which is significantly more accurate than orthographic methods. In perspective depth, each light ray is traced to the exact pixel from the object its coming from; in orthographic depth, on the other hand, light-rays are assumed to be coming from \textit{infinity} (see Fig.~\ref{fig:prespvortho}). In real-world datasets, e.g., NYUv2 \cite{Silberman:ECCV12} and DIODE \cite{diode_dataset}, depth is registered based on orthographic projection because of physical limitations in the sensor.

\vspace{0.4em}
\noindent\textbf{Surface normals:} We estimate the normal vector w.r.t to each 3D surface, then color each pixel to indicate the vector's direction. We use 6 main colors to show 6 axis of direction (positive and negative xyz, as shown in Fig.~\ref{fig:envexamples}).

\vspace{0.4em}
\noindent\textbf{Semantic segmentation:} In UE4, it is easy to map visible pixels to their corresponding 3D objects. Our Blueprint script uses this mapping to overlay pixel-perfect semantic labels on the various objects in the scene (e.g., couch, table, lamp, etc.). 

\begin{figure}[htbp]
\centerline{\includegraphics[width = 0.73\textwidth]{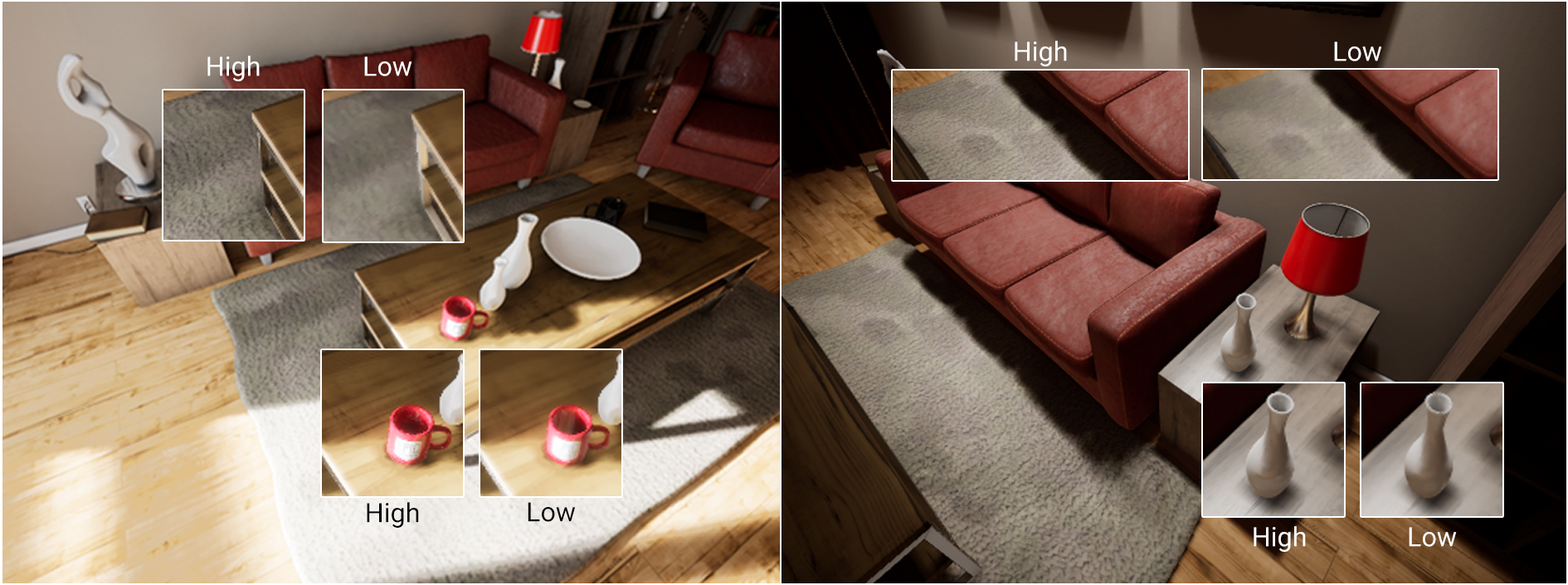}}
\caption{\textbf{Fidelity Comparison:} Left: Day(High Fidelity), Right: Night(High Fidelity). Each image snippet of Low Fidelity indicates the difference in Texture resolution, Reflections quality, Render Scaling and Shadow quality. The amount of change in each of these settings is customizable through AIP's Core.}
\label{fig:fidelityshow}
\vspace{-1em}
\end{figure}

\subsection{Data ablation controls}
Similar to the ground-truth, we use Blueprint to dynamically alter properties of the environment. We can access and isolate specific properties in different objects. For example, we can isolate metallic objects or rough surfaces with a pixel-perfect binary ground truth. We can also change the fidelity of reflections, lighting, mesh level of detail (LOD), render resolution (either localized to an object or globally), anti-aliasing algorithms (or toggle on and off), or render scaling. Figure~\ref{fig:fidelityshow} illustrates the same scene rendered under different fidelity settings. Our scripts are reusable, in the sense that they do not require adaptation to other environments and are also easily portable to other UE projects.

\subsection{User interface}
The AIP Core can be opened as a project in UE4, giving access to all its assets and scripts. Alternatively, we provide a pre-compiled version which can be run as an independent program. AIP has intuitive user menus and keyboard shortcuts. Our Python Probe script uses the latter to collect data (see Sec.~\ref{sec:experiments} for details). 

\vspace{-2em}
\begin{table}[h]
\centering
\caption{Scenarios used in experiments$^{\mathrm{a}}$}
\label{tab:scenarios}
\resizebox{0.73\linewidth}{!}{%
\begin{tabular}{|l|l|l|c|} 
\hline
\textbf{Default}                 & \multicolumn{3}{c|}{\textbf{Settings} }                                                          \\ 
\cline{2-4}
 \textbf{Maps}                   & \textbf{\textit{Lighting}} & \textbf{\textit{Fidelity}} & \textbf{\textit{Anti-Aliasing}}        \\ 
\hline
Brown Room                       & Day                        & High                       & Temporal AA                            \\ 
\hline
Brown Room                       & Night                      & High                       & Temporal AA                            \\ 
\hline
Brown Room                       & Day                        & Low                        & Temporal AA                            \\ 
\hline
Brown Room                       & Night                      & Low                        & Temporal AA                            \\ 
\hline
Blue Room                        & Day                        & High                       & Temporal AA                            \\ 
\hline
Blue Room                        & Night                      & High                       & Temporal AA                            \\ 
\hline
Blue Room                        & Day                        & Low                        & Temporal AA                            \\ 
\hline
Blue Room                        & Night                      & Low                        & Temporal AA                            \\ 
\hline
Abstract Shapes                  & Day                        & High                       & Temporal AA                            \\ 
\hline
Unlit$^{\mathrm{b}}$ Brown Room  & N/A                        & High                       & Temporal AA                            \\ 
\hline
Unlit Blue Room                  & N/A                        & High                       & Temporal AA                            \\ 
\hline
\multicolumn{4}{l}{$^{\mathrm{a}}$shows settings used, not indicative of all settings available.$^{\mathrm{b}}$diffuse shading.}   
\end{tabular}
}
\end{table}
\vspace{-2em}


\begin{figure}[t]
\centerline{\includegraphics[width = 1\textwidth]{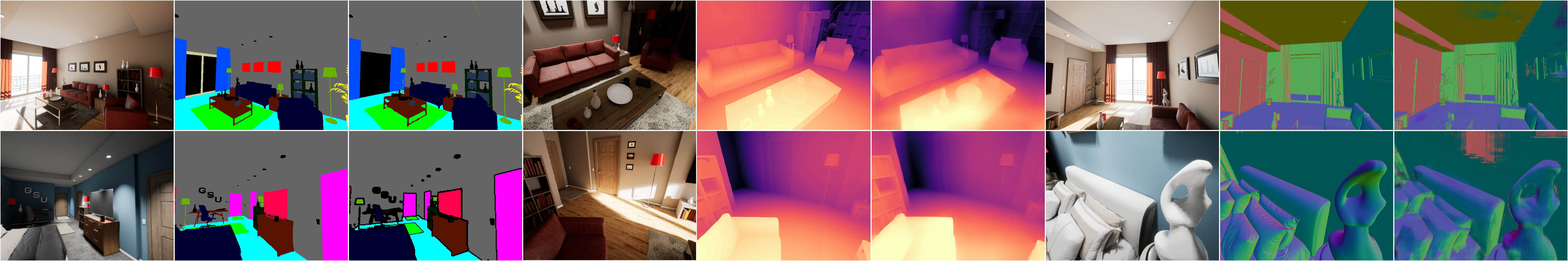}}
\caption{\textbf{Sample results:} Sample images, ground truth, and predictions for semantic segmentation (first three columns), depth estimation (middle columns), and surface normal estimation (last three columns). Figure best viewed onscreen.}
\label{fig:fullshowcase}
\vspace{-1.25em}
\end{figure}

\section{Experiments and Results}
\label{sec:experiments}
We carried out multiple experiments to validate the usefulness of our proposed system. Specifically, we tested AIP in two ways. First, we verified its viability as a data ablation tool. As we detail below, we captured the same images under different fidelity and lighting settings (which we refer to as a \textit{scenario}), then trained deep neural networks on each scenario to assess the impact of the various environmental features. We carried out both same- and cross-scenario testing (e.g., a Brown/Day/High network on Brown/Night/High). Table~\ref{tab:scenarios} summarizes the scenarios used. For each scenario, we tested our networks on (1) monocular depth and (2) surface normal estimation, as well as (3) semantic segmentation. 

Second, to validate that our virtual data is realistic enough, we tested networks trained with AIP on real-world depth-estimation datasets, achieving results comparable to training on real data alone. Below, we first detail our experimental setup, then discuss each experiment.

\subsection{Experimental Setup}

\noindent{\textbf{Hardware:}} We conducted all our experiments in a Dell Precision 7920R server with two Intel Xeon Silver 4110 CPUs, two GeForce GTX 1080 Ti graphics cards, and 128 GBs of RAM.
\vspace{0.4em}

\noindent\textbf{Image acquisition:} Our Probe script can control the viewpoint by simulating keystrokes. It can move and look freely (yaw and pitch) in the environment. Probe can also send specific commands and can gather images with high overlap (in groups) or low overlap (completely random). Probe's step size, look sensitivity, randomness of image acquisition (group capture), and number of images to gather are all customizable and can be saved for reproduction across all different scenarios. For our depth estimation experiments, we randomly collected 8265, 640$\times$480 synthetic color images. We collected the same images, by replicating the same camera positions and rotations, across different lighting and fidelity scenarios (Tbl.~\ref{tab:depth_ablation}). We split these images into 80\% for training, and 20\% for testing. Similarly, for semantic segmentation and surface normal estimation, we gathered 3000 images for each scenario and split in the same ratio.
\vspace{0.4em}




\noindent{\textbf{Deep neural networks:}} We used the encoder-decoder architecture, and loss function from \cite{Silberman:ECCV12} for depth estimation, and an implementation of U-net \cite{Ronneberger2015UNetCN} from \cite{DBLP:journals/corr/abs-1903-07803_deepdyn} for surface normal estimation and semantic segmentation. We use smooth L1 loss function for Surface Normal Estimation, and Cross-Entropy loss for segmentation task. We use a \textit{mini-batch size} of 16, \textit{learning rate} of  0.001, and trained for 51 \textit{epochs} for all experiments. 



\begin{table}[t]
\centering
\caption{\textbf{Depth estimation:} Data ablation test results. Metrics are threshold accuracy ($\delta_i < 1.25^i$), average relative error (REL), root mean squared error (RMS), and average (log10) error. Arrows indicate if higher or lower values are better. For space, we included only some of the conducted experiments; results shown are indicative of the behavior of the trained models in other scenarios. \newline \textbf{SC:} Sanity Check. \textbf{L:} Change in Lighting. \textbf{M:} Change in Maps. \textbf{F:} Positive Change in Fidelity}
\label{tab:depth_ablation}
\resizebox{\textwidth}{!}{%
\begin{tabular}{|l|l|l|c|c|c|c|c|c|}
\hline
\textbf{Training Scenario /  Fidelity} & \textbf{Test / Fidelity} & \textbf{Goal} & \textbf{$\delta_1\uparrow$} & \textbf{$\delta_2\uparrow$} & \textbf{$\delta_3\uparrow$} & \textbf{REL$\downarrow$} & \textbf{RMS$\downarrow$} & \textbf{log10$\downarrow$} \\ \hline
Brown / Day / High                             & Brown/ Day / High                     & \textbf{SC}   & 0.7992      & 0.9113      & 0.9474      & 0.1426       & 0.0278       & 0.0740         \\ \hline
Blue / Day / Low                & Blue / Day / Low        & \textbf{SC}   & 0.7609      & 0.8980      & 0.9278      & 0.1643       & 0.0366       & 0.0858         \\ \hline
Brown / Night / High                           & Brown / Night / High                   & \textbf{SC}   & 0.8333      & 0.9248      & 0.9509      & 0.1327       & 0.0278       & 0.0689         \\ \hline
Brown/ Day / Low                              & Brown/ Day / Low                      & \textbf{SC}   & 0.7719      & 0.8945      & 0.9388      & 0.1544       & 0.0289       & 0.0798         \\ \hline
Brown/ Day / High                             & Brown/ Night / High                   & \textbf{L}    & 0.7616      & 0.8928      & 0.9315      & 0.1711       & 0.0398       & 0.0875         \\ \hline
Brown/ Night / High                           & Brown/ Day / High                     & \textbf{L}    & 0.7366      & 0.8942      & 0.9420      & 0.1904       & 0.0351       & 0.0939         \\ \hline
Blue / Day / Low                & Blue / Day / High       & \textbf{F}    & 0.7817      & 0.9062      & 0.9329      & 0.1587       & 0.0370       & 0.0822         \\ \hline
Brown/ Day / Low                              & Brown/ Day / High                     & \textbf{F}    & 0.8010      & 0.9113      & 0.9475      & 0.1426       & 0.0273       & 0.0731         \\ \hline
Brown/ Night / High                           & Blue / Night / High     & \textbf{M}    & 0.5959      & 0.8632      & 0.9079      & 0.3415       & 0.0671       & 0.1193         \\ \hline
Brown / Day / High                             & Blue Day / High       & \textbf{M}    & 0.6420      & 0.8528      & 0.9223      & 0.2220       & 0.0433       & 0.1067         \\ \hline
\end{tabular}%
}
\vspace{-1.5em}
\end{table}

\begin{table}
\centering
\caption{\textbf{Depth estimation:} Results on real-world datasets.}
\label{tab:extradepth}
\resizebox{\linewidth}{!}{%
\begin{tabular}{|l|l|c|c|c|c|c|c|} 
\hline
 \textbf{Train / Fidelity}  & \textbf{Test}  & \multicolumn{1}{l|}{\textbf{$\delta_1\uparrow$} } & \multicolumn{1}{l|}{\textbf{$\delta_2\uparrow$} } & \multicolumn{1}{l|}{\textbf{$\delta_3\uparrow$} } & \multicolumn{1}{l|}{\textbf{REL$\downarrow$} } & \multicolumn{1}{l|}{\textbf{RMS$\downarrow$} } & \multicolumn{1}{l|}{\textbf{log10$\downarrow$} }  \\ 
\hline
Brown / Day / High                  & NYUv2                         & 0.3666                                            & 0.6012                                            & 0.7586                                            & 0.5044                                         & 0.2014                                         & 0.1938                                            \\ 
\hline
Brown / Day / Low                   & NYUv2                         & 0.3720                                            & 0.6062                                            & 0.7627                                            & 0.5010                                         & 0.2010                                         & 0.1921                                            \\ 
\hline
Brown / Night / High                & DIODE                         & 0.3563                                            & 0.5948                                            & 0.7945                                            & 0.7659                                         & 3.6897                                         & 0.2148                                            \\ 
\hline
Brown / Night / Low                 & DIODE                         & 0.3163                                            & 0.5647                                            & 0.7345                                            & 0.7743                                         & 3.7898                                         & 0.2149                                            \\ 
\hline
Brown / Night / High                & DIODE - Filtered              & 0.6546                                            & 0.7725                                            & 0.8371                                            & 0.6608                                         & 2.9765                                         & 0.1458                                            \\ 
\hline
Brown / Day / High                  & NYUv2 - Filtered              & 0.5996                                            & 0.8405                                            & 0.9308                                            & 0.2835                                         & 0.1232                                         & 0.1054                                            \\ 
\hline
DIODE/Indoor\cite{diodepaper}                & DIODE/Indoor                  & 0.4919                                            & 0.7159                                            & 0.8256                                            & 0.3306                                         & 1.6948                                         & 0.1775                                            \\ 
\hline
NYUv2\cite{Alhashim2018_tf_impl}                       & NYUv2                         & 0.895                                             & 0.980                                             & 0.9960                                            & 0.1030                                         & 0.390                                          & 0.0430                                            \\ 
\hline
NYUv2\cite{Alhashim2018_tf_impl}                     & DIODE/Indoor                  & 0.2869                                            & 0.5097                                            & 0.6730                                            & 0.6599                                         & 2.8854                                         & 0.2573                                            \\
\hline
\end{tabular}
}
\vspace{-1.5em}
\end{table}

\subsection{Monocular depth estimation experiments}
\noindent\textbf{Data ablation:} Table~\ref{tab:depth_ablation} shows a representative sample of the data ablation experiments we conducted using our depth ground truth. For these experiments, we initialized our deep networks using the weights from a network trained on NYUv2. For evaluation, we used the same metrics as those used in \cite{DBLP:journals/corr/EigenPF14_sing_img_depth}: average relative error (REL), root mean squared error (RMS), average log10 error, and threshold accuracy ($\delta_i < 1.25^i$ for $i = [1,2,3]$). As we discuss further in Sec.~\ref{sec6:Discussion}, models trained in higher fidelity data generally tend to yield higher scores, even on lower-fidelity scenarios.
\vspace{0.5em}


\noindent\textbf{Real-world validation:} To demonstrate the transferability of learned features from a synthetic dataset, we tested our best-performing models on the real-world DIODE and NYUv2 datasets. In addition to the full test set, we also evaluated our networks on a filtered subset that only contained scenes structurally similar to our virtual environments, i.e., indoor scenes of a living room, with objects such as couches, beds, TVs, etc. As Tbl.~\ref{tab:extradepth} shows, our high-fidelity trained model had better threshold accuracy on DIODE than a model trained only on NYUv2 \cite{diodepaper}, confirming that the features learned on our environments are transferable to real-world data. In addition, our model trained on Night lighting, high-fidelity settings achieved 31\% $\delta_1$ vs 28\% $\delta_1$ of NYUv2 model | 59\% $\delta_2$ vs 50\% $\delta_2$ of NYUv2 model | 79.4\% $\delta_3$ vs 67.3\% of  $\delta_3$ of NYUv2 model. These results further confirm that our photo-realistic data can match and even exceed real-life training. Furthermore, these models achieved a much higher score in our filtered test set, suggesting that depth estimation is more sensitive to the structure of the input image than to lighting or fidelity. We also believe our models would have performed even better had DIODE used perspective depth (Fig.~\ref{fig:prespvortho}).





\begin{table}[t]
\centering
\caption{\textbf{Surface normal estimation:} Metrics are percentage of pixels that differ by \ang{11.5}, \ang{22.5}, and \ang{30} from the true normal, and mean and median errors. Mean and median are higher than \cite{DBLP:journals/corr/abs-1904-03405_normal} because our loss function did not implement hybrid measures to reduce them. This wasn't necessary since our ground-truth data does not suffer from the problem mentioned in \cite{DBLP:journals/corr/abs-1904-03405_normal}.  \newline \textbf{SC:} Sanity Check. \textbf{L:} Change in Lighting. \textbf{M:} Change in Maps. \textbf{F:} Positive Change in Fidelity}
\label{tab:normal_results}
\resizebox{\textwidth}{!}{%
\begin{tabular}{|l|l|c|c|c|c|c|c|}
\hline
\textbf{Scenario / Fidelity} & \textbf{Test / Fidelity} & \multicolumn{1}{l|}{\textbf{Goal}} & \multicolumn{1}{l|}{\textbf{$\ang{11.5}\uparrow$}} & \multicolumn{1}{l|}{\textbf{$\ang{22.5}\uparrow$}} & \multicolumn{1}{l|}{\textbf{$\ang{30}\uparrow$}} & \multicolumn{1}{l|}{\textbf{Mean}$\downarrow$} & \multicolumn{1}{l|}{\textbf{Median}$\downarrow$} \\ \hline
Brown / Day / High           & Brown / Day / High       & \textbf{SC}                        & 0.9014                           & 0.9566                           & 0.9727                           & 24.4575                            & 88.2878                              \\ \hline
Blue / Day / Low             & Blue / Day / Low         & \textbf{SC}                        & 0.9274                           & 0.9746                           & 0.989                            & 30.5607                            & 94.9516                              \\ \hline
Blue / Night / High          & Blue / Night / High      & \textbf{SC}                        & 0.865                            & 0.9224                           & 0.9401                           & 28.2409                            & 69.2181                              \\ \hline
Brown / Day / Low            & Brown / Day / Low        & \textbf{SC}                        & 0.8883                           & 0.9443                           & 0.961                            & 25.3718                            & 81.4871                              \\ \hline
Brown / Day / High           & Brown / Night / High     & \textbf{L}                         & 0.052145                         & 0.2238                           & 0.3464                           & 106.70                             & 121.26                               \\ \hline
Brown / Night / High         & Brown / Day / High       & \textbf{L}                         & 0.050291                         & 0.2135                           & 0.4253                           & 115.82                             & 119.86                               \\ \hline
Blue / Day / Low             & Blue / Day / High        & \textbf{F}                         & 0.195269                         & 0.2683                           & 0.3015                           & 97.832                             & 113.57                               \\ \hline
Brown / Day / Low            & Brown / Day / High       & \textbf{F}                         & 0.028247                         & 0.2102                           & 0.368                            & 109.14                             & 118.08                               \\ \hline
\end{tabular}%
}
\vspace{-1em}
\end{table}

\vspace{-0.5em}
\subsection{Surface normal estimation experiments}
We carried out a similar set of data ablation experiments as above, but using surface normal data as the ground truth. Here, we trained each model from scratch, i.e., without pre-trained weights, and used the same evaluation metrics as in \cite{DBLP:journals/corr/abs-1904-03405_normal}: mean (average L1 loss), median (average L2 loss), and percentage of pixels that differ by \ang{11.5}, \ang{22.5}, and \ang{30} relative to the true surface normal. Surface normal estimation is a promising use case for AIP because it is very challenging to capture surface normal information for real scenes. One needs expensive equipment to measure the angles, and these sensors are extremely hard to calibrate. As Tbl.~\ref{tab:normal_results} shows, we can successfully train deep networks using AIP (see Fig.~\ref{fig:fullshowcase}). Overall, surface normal models seem to be less sensitive to photo-realistic features and higher fidelity settings compared to depth estimation or segmentation. Models trained on high fidelity settings perform 2$\%$ better than ones trained on low fidelity, a point we discuss further in Sec.~\ref{sec6:Discussion}.

\subsection{Semantic segmentation experiments}
Semantic segmentation involves assigning a class label to every pixel on the image. The built-in environments in AIP have fifteen classes, all of which corresponds to regular household objects, e.g., \textit{wall}, \textit{couch}, \textit{table}, \textit{TV}, \textit{plant}, etc. We use a label of \textit{other} for miscellaneous items. As with the surface normals, we trained different networks from scratch on each scenario. We used mean intersection over union (IOU) of all classes as our evaluation metric. As we can see in Tbl.~\ref{tab:segmentation}, model performance is directly linked to a scenario's fidelity (see Fig.~\ref{fig:fullshowcase}). Semantic segmentation seems to depend heavily on the render scaling and resolution. At lower settings, borders of the objects are blurry, as is their texture. This causes the model to label them as \textit{other} since it cannot surely ascertain their object class, thus lowering the global IOU (see Fig.~\ref{fig:segablation} for an example).

\begin{table}[t]
\caption{\textbf{Semantic segmentation:} Mean intersection over union (IOU) of all classes for different scenarios. Higher values are better. \newline \textbf{SC:} Sanity Check. \textbf{L:} Change in Lighting. \textbf{M:} Change in Maps. \textbf{F:} Positive Change in Fidelity}
\centering
\resizebox{0.73\textwidth}{!}{%
\begin{tabular}{|l|l|c|c|}
\hline
\textbf{Scenario / Fidelity} & \textbf{Test / Fidelity}        & \multicolumn{1}{l|}{\textbf{Goal}} & \multicolumn{1}{l|}{\textbf{Global IOU$\uparrow$}} \\ \hline
Brown / Day / High           & Brown / Day / High   & \textbf{SC}                        & 0.8984                                   \\ \hline
Blue / Day / Low             & Blue / Day / Low     & \textbf{SC}                        & 0.4119                                   \\ \hline
Blue / Night / High          & Blue / Night / High  & \textbf{SC}                        & 0.8335                                   \\ \hline
Brown / Day / Low            & Brown / Day / Low    & \textbf{SC}                        & 0.4714                                   \\ \hline
Brown / Day / High           & Brown / Night / High & \textbf{L}                         & 0.6932                                   \\ \hline
Brown / Night / High         & Brown / Day / High   & \textbf{L}                         & 0.6418                                   \\ \hline
Blue / Day / Low             & Blue / Day / High    & \textbf{F}                         & 0.3862                                   \\ \hline
Brown / Day / Low            & Brown / Day / High   & \textbf{F}                         & 0.4188                                   \\ \hline
\end{tabular}%
}
\label{tab:segmentation}
\end{table}
\vspace{-1em}


\section{Discussion}
\label{sec6:Discussion}
Below, we discuss some insights from our data ablation experiments that serve as examples of the kind of analyses that AIP makes possible.
\vspace{0.5em}



\begin{figure}[t]
\centerline{\includegraphics[width = 0.63\textwidth]{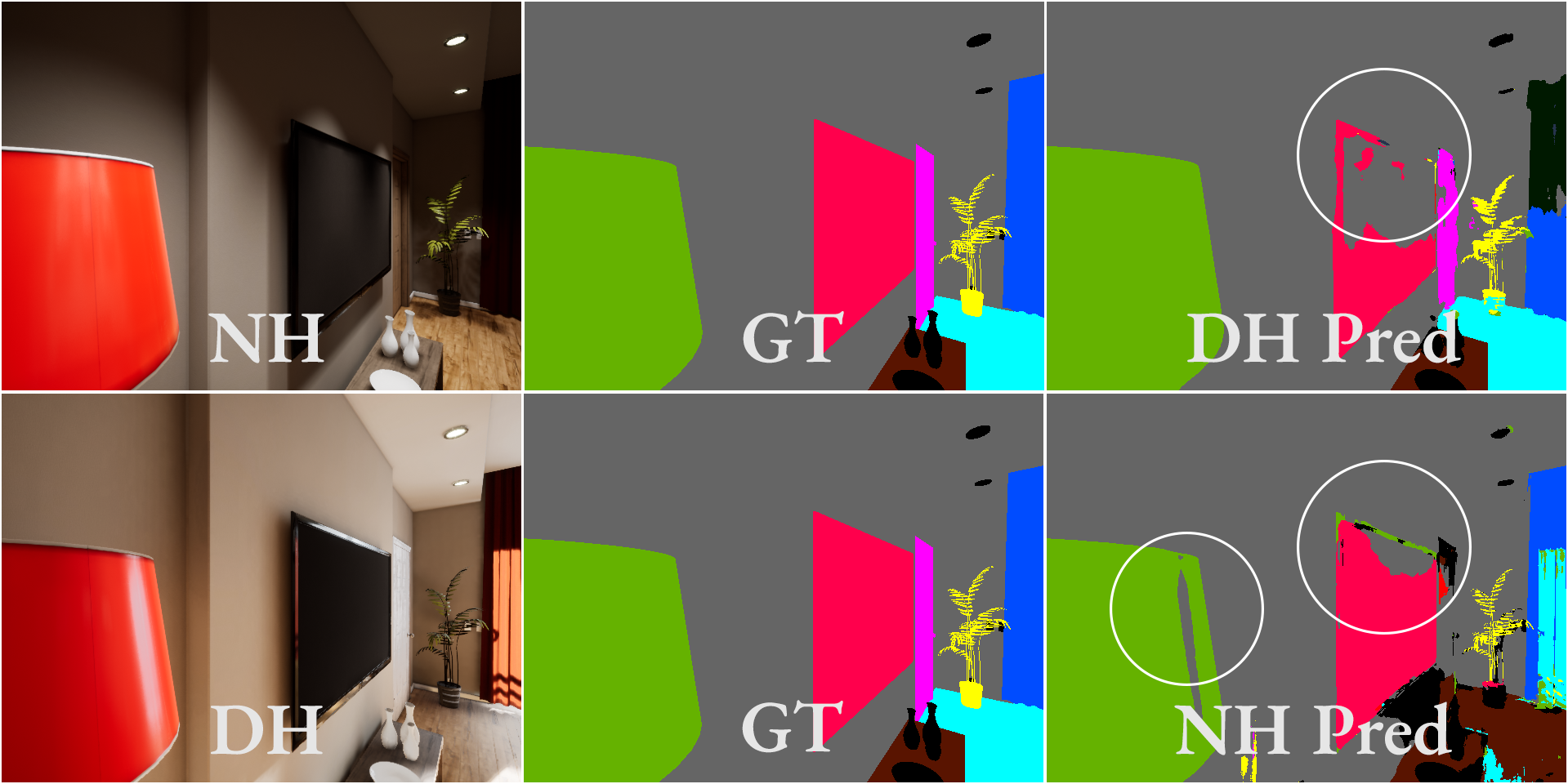}}
\caption{\textbf{Semantic segmentation:} (Image, Ground Truth, Prediction). Top: A model trained on Brown Day High (DH) images segmenting a Brown Night High (NH) image. Bottom: a model trained on Brown Night High tested on Brown Day High. Note the impact of lighting on the final result.}
\label{fig:segablation}
\end{figure}

\noindent\textbf{Sensitivity to lighting:} Changes in lighting are a result of the environment, so they cannot be "fixed" by a better acquisition device. As such, a general-purpose model should be robust to them. However, objects can appear in drastically different ways under different lighting conditions, which did affect performance across all experiments. More specifically, segmentation models are particularly sensitive to differences in lighting. In Fig.~\ref{fig:segablation} both models labeled the top part of the \textit{TV} as \textit{Wall} since they have almost the same color. However, the model trained on a Day setting was much less accurate on the Night image than its counterpart, presumably because the Night setting is darker overall and has more pronounced reflections. The opposite effect is visible in the reverse case (bottom Fig.~\ref{fig:segablation}), where the reflection in the lamp confused the model because that level of reflection from sunlight does not exist in the Night lighting.

Our surface normal models are also sensitive to changes in lighting. However, for depth estimation, performance drops only slightly when the lighting is changed, suggesting that local contrast is less important for this problem.



\vspace{0.5em}
\noindent\textbf{The impact of fidelity on surface normals vs. segmentation:} Semantic segmentation is very sensitive to changes in fidelity. When objects are blurred due to lower rendering resolution and lower texture clarity, the model appears to be indecisive about picking an object's class in its border regions. As shown in Fig.~\ref{fig:fullshowcase}, we see that the model incorrectly classified border regions as \textit{Other}.

In contrast, surface normal estimation is more robust to these kinds of changes. This difference between these two problems highlights the importance of using data ablation tools. Previous studies, e.g., \cite{7926706_photo_realism, Photorealism2, photorealism3}, mainly focus on the effects of fidelity on their segmentation experiments. Our findings with surface normals, on the other hand, suggest that fidelity as a general feature of the image might not be enough to draw conclusions about the quality of the data. AIP's tools allow us to study other aspects of data, such as texture, structure complexity, lighting and more.


\vspace{0.5em}
\noindent\textbf{Perspective vs orthographic depth:} Orthographic depth projection is when light-rays coming to the camera are assumed to be coming from \textit{infinity}. In calculating the depth ground-truth, this simplification introduces errors to the measurement. We have seen the effects of this assumption on the NYUv2 and DIODE dataset (Fig.~\ref{fig:prespvortho}). Specifically, our models' performance on DIODE was lower in part due to them being trained on perspective depth, which is different from the GT used in DIODE. Although orthographic measurements are currently widely used, we argue that perspective depth, which AIP supports, is the \textit{correct} way to measure depth.


\vspace{0.5em}
\noindent\textbf{Impact of fidelity on depth estimation:}
Generally, the performance of models trained on higher fidelity settings are better than those trained in lower fidelity settings (Table \ref{tab:depth_ablation}). However, one exception is when the lower fidelity setting in training better matches the features of the target domain. In Tbl.~\ref{tab:extradepth}, our low fidelity model does slightly better on NYUv2 than the high-fidelity one. We argue this is due to the blur present in NYUv2, which is also present in our low fidelity settings training set due to its lower render settings, making them visually similar. The DIODE dataset, on the other hand, is much more recent, so the depth ground truth was measured with a more accurate sensor. Due to the lack of blur and fuzz on the ground-truth, we did not observe the same kind of performance gain on this dataset.

\section{Conclusion \& Future work}
\label{sec:conclusions}
\vspace{-0.5em}
In this work, we presented AI Playground, a data creation and ablation tool for machine learning. Using AIP, we generated different image datasets and conducted experiments that are nearly impossible with real data, thus demonstrating that AIP is a viable tool for data ablation studies in computer vision. We also verified that our high-fidelity trained models can match or exceed the scores achieved by training with real-data. As suggested by other studies  \cite{7926706_photo_realism,Photorealism2,photorealism3, datafromgames}, we found that higher-fidelity data is linked to better performance in segmentation, but we also found that sensitivity to scene structure, fidelity and lighting scenario of training data varies from task to task. For example, our surface normal and depth estimation models were not as sensitive to fidelity as our segmentation models were. AIP enables us to change individual features, e.g., quality of shadows, quality of reflections, quality of lighting or resolution of textures, and assess their impact on different models based on the current task. More generally, AIP can help researchers find sensitive points in their models and aid them in creating high-quality data for training neural networks for a specific computer vision task.

We are currently working to add more environments to AIP to widen its usage range. These environments include more indoor scenes, outdoor scenes and fully interactive environments allowing individual interaction with objects. Additionally, we'll be providing support for reinforcement learning studies and real-time ray-tracing. There are still many other possible experiments that remain to be explored. For example, UE4 allows the fast change of lighting profile by using HDRI maps. This opens the possibility of adding more specific lighting scenarios like rainy, overcast and foggy. In our future updates, we'll be adding support to introduce intentional camera artifacts such as chromatic aberration, penumbra, lens flares and distortions to help study the effects of using small sensors in capturing data. This is especially useful in robotics since consumer-grade robots rarely come with expensive capture equipment; fine-tuning training to the exact specifications of the camera is a very exciting avenue for future work. Furthermore, we are refining our ground-truth options, including removing texture and changing colors and properties of shaders. These enhancements will enable us to manipulate the scene even further, e.g., changing the pattern in a fabric or changing smoothness of a stone. We believe that AIP will open new and exciting avenues in synthetic data and machine learning.

\bibliographystyle{splncs04}
\bibliography{bibliography}

\end{document}